\documentclass[runningheads]{llncs}

\usepackage{accv}

\usepackage{accvabbrv}

\usepackage{graphicx}
\usepackage{booktabs}
\usepackage{array}
\usepackage{tabularx}
\usepackage[accsupp]{axessibility}  

\usepackage[pagebackref,breaklinks,colorlinks,citecolor=accvblue]{hyperref}

\usepackage{orcidlink}

\begin{document}

\title{Validating FKG.in: Soundness Assessment in LLM-Augmented Indian Food Knowledge} 

\titlerunning{Validating FKG.in}

\author{Saransh Kumar Gupta\inst{1}\orcidlink{0009-0000-5887-2301}\thanks{Corresponding Author.} \and
Armaan Shah\inst{1}\thanks{S.K. Gupta and A. Shah contributed equally.} \and
Lipika Dey\inst{1}\orcidlink{0000-0003-3831-5545} \and
Partha Pratim Das\inst{1}\orcidlink{0000-0003-1435-6051} \and
Ramesh Jain\inst{2}\orcidlink{0000-0003-2373-4966}}

\authorrunning{S.K. Gupta et al.}


\institute{Ashoka University, India\\
\email{\{saransh.gupta,lipika.dey,partha.das\}@ashoka.edu.in}\\
\email{armaan.shah@alumni.ashoka.edu.in}\\
\url{https://www.ashoka.edu.in/} \and
Institute for Future Health, UC Irvine, USA\\
\email{jain\@ics.uci.edu}\\
\url{https://futurehealth.uci.edu/}}

\maketitle

\begin{abstract}
  The online culinary ecosystem is increasingly populated by recipe content generated, modified, or summarized by LLMs. Although often convincing, such outputs may contain hallucinated ingredients, misrepresented quantities, or culturally implausible combinations, limiting their suitability for downstream applications and knowledge graph construction. In this paper, we present a semi-automated soundness assessment workflow for validating structured recipe data extracted from informal culinary sources and augmented by LLMs. Developed as part of FKG.in, a knowledge graph of Indian food, the pipeline identifies and addresses common failure modes, including structural inconsistencies, semantic and logical incoherence, and deviations from the source text, through a multi-stage process combining formal grammars, vocabulary-based checks, statistical heuristics, Set Transformer-based coherence modeling, and retrieval-based verification. Although evaluated on Indian recipes, the proposed methods are applicable to broader multilingual and multicultural culinary domains. We provide a practical, auditable, and application-agnostic approach to improving the reliability of machine-readable food knowledge infrastructures in the era of LLM-generated content.
  \keywords{Food Computing \and Knowledge Engineering \and Information Validation \and Large Language Models \and Semantic Assessment \and
  Indian Food}
\end{abstract}

\section{Introduction}

\label{sec:introduction}

Indian cuisine is one of the world's most diverse culinary traditions, yet structured representations of Indian food knowledge remain scarce and fragmented across cookbooks, recipe blogs, food wikis, Kaggle datasets, community forums, YouTube videos, and food journals. Although these sources are often informal, noisy, and inconsistent, they contain valuable culturally grounded culinary knowledge. To address this gap, we developed FKG.in \cite{Gupta2024FKG}, a structured, machine-readable knowledge graph constructed from diverse Indian culinary sources.

Food data from such sources are increasingly used in applications including personalized diet recommendations, nutrition tracking, recipe generation, ingredient substitution, and culinary knowledge graph construction. However, India's linguistic, regional, and culinary diversity creates significant challenges for reliable food information extraction and named entity recognition (NER). Ingredient and dish names vary across languages and regions, quantity descriptions are often imprecise or culturally specific, and recipe formats differ widely in structure, granularity, and linguistic expression \cite{Gupta2025FKG}. These variations make it challenging to accurately identify, extract, and standardize food-related entities across diverse Indian culinary data at scale.

Large Language Models (LLMs) have helped address many of these challenges by enabling structured information extraction from noisy, multilingual, and highly variable recipe sources. Yet validating LLM-generated data remains a critical bottleneck. Their outputs can contain factual inaccuracies, logical inconsistencies, omissions, semantic distortions, and hallucinations that are not always immediately apparent but can significantly affect downstream applications. In the food domain, even seemingly minor errors can alter the nutritional, functional, or cultural characteristics of a recipe. For example, an LLM may infer an ingredient commonly associated with a dish but absent from the source, producing an output that appears plausible but lacks source fidelity. Such errors can propagate to applications including nutritional analysis, dietary recommendations, recipe retrieval, and knowledge graph construction, and food safety.

In this paper, we present a semi-automated workflow for validating structured information extracted by LLMs as part of the FKG.in knowledge curation process. We assess factual correctness, internal consistency, semantic plausibility, and fidelity to the source text using rule-based checks, statistical heuristics, semantic consistency modeling, retrieval-based verification, and human-in-the-loop review. The workflow identifies and addresses inconsistencies before LLM-extracted information is incorporated into FKG.in and related food knowledge infrastructures. The remainder of the paper reviews related work, examines challenges in validating LLM-extracted food knowledge, presents the proposed soundness assessment workflow, reports evaluation results, and discusses limitations and future directions.

\section{Related Work}
\label{sec:related_work}

The increasing use of LLMs for knowledge generation has highlighted their tendency to hallucinate, producing fluent but factually incorrect outputs. Research on addressing this problem broadly falls into two areas: general-purpose hallucination detection and verification, and domain-specific validation, including recent work in the food domain.

General-purpose factuality research has explored a range of verification strategies, including claim decomposition and entailment-based verification \cite{glover-etal-2022-revisiting,wanner2024dndscore}, retrieval-verifier pipelines and retrieval-aware training \cite{zhang2023relevance,zheng2024evidence}, direct evidence generation \cite{chen2022gere}, interpretable reasoning and proof-based verification \cite{chen2022loren,krishna2022proofver}, and knowledge graph-oriented verification \cite{amaral2024prove, venkat2026}. Related work has also examined the construction and evaluation of LLM-generated knowledge graphs using semantic and LLM-based assessment methods \cite{nechakhin2024evaluating}. However, many of these approaches assume well-formed natural language claims and evidence, making their application to noisy, heterogeneous culinary data non-trivial.

In the food domain, efforts to validate LLM-generated content remain relatively limited. Existing studies have examined hallucination risks in nutrition- and ingredient-related outputs \cite{tian2025large,pitsilou2024using,ravichander2025halogen}, while other work has focused on culinary language understanding, recipe generation, and ingredient substitution reasoning through domain adaptation and benchmark development \cite{hua2024nutribench,thornton2024nutritional}. Food-specific retrieval-augmented generation, domain-adapted culinary LLMs, and graph-based personalization frameworks have also been explored \cite{akilesh2025graph,zhou2024foodsky,vij2025fine}. Structured knowledge representations have further been proposed to improve traceability and reasoning in LLM-driven systems \cite{erickson2025llm}.

\section{Challenges in Validating LLM-Extracted Food Knowledge}
\label{sec:challenges}

Despite advances in LLM factuality verification and food-related applications, validation frameworks for LLM-extracted food knowledge remain underdeveloped. Existing approaches primarily focus on generation quality, personalization, or domain adaptation rather than systematically assessing the soundness of extracted recipe data. We identify three core failure modes encountered during FKG.in knowledge curation and describe corresponding validation strategies below, drawing on examples from our experimental workflow. The validation strategies continue to evolve as the system matures.



\begin{enumerate}
    \item \textbf{Structural and Numerical Inconsistency}: This failure mode arises when LLM outputs violate the expected ontology schema or produce numerically implausible or ambiguous quantities. Indian recipe sources vary considerably in how ingredients and quantities are represented, ranging from nested ingredient hierarchies to flat enumerations and informal measurements such as \textit{katori} (small bowl), handful, or pinch, making normalization difficult. Examples include a single-serving \textit{dal} (lentil dish) recipe incorrectly specifying `5 kilograms of lentils', the ambiguous quantity `2 cups of rice' without contextual information about serving size or yield, malformed key-value pairs such as `2 grams: salt' instead of `salt: 2 grams', and unit-ingredient mismatches such as `3 tablespoons of cardamom pods' in a dessert intended for two servings instead of a more typical quantity such as `3 cardamom pods.' We address these issues using schema validation, formal grammars, rule-based checks, ingredient-quantity distribution models, and heuristics for normalizing ambiguous or culturally specific measurements.
    
    \item \textbf{Semantic and Logical Incoherence}: Even when ingredient entries are structurally valid and numerically reasonable, they may be semantically implausible, chemically unsound, or culturally inappropriate. Such errors can arise when LLMs generalize across noisy and heterogeneous culinary data, producing substitutions, hallucinations, or combinations that violate culinary logic without triggering structural or quantity constraints. For example, an LLM may substitute \textit{maida} (refined wheat flour) for \textit{besan} (gram flour) in dishes such as \textit{kadhi} (spiced yogurt-based stew) or \textit{cheela} (savory pancake), mistaking one for the other because both are commonly described as flour and share usage contexts (e.g., mixing into batters) despite their distinct functional and flavor roles. Similarly, it may hallucinate tuna in a vegetarian dish such as \textit{paneer tikka} (grilled Indian cottage cheese), violating dietary expectations, or combine tempering of mustard oil and curry leaves with a Gujarati sweet dish such as \textit{shrikhand} (sweetened strained yogurt dessert), producing a pairing that conflicts with established culinary conventions. We address these issues using vocabulary-based checks, ingredient co-occurrence statistics, and a Set Transformer model trained on heuristic and semantic compatibility signals to assess contextual fit. Unfamiliar or region-specific ingredients outside known vocabulary and co-occurrence distributions are flagged for human review to distinguish genuine culinary variation and previously unseen vernacular terms from hallucinated or erroneous entities.

    
    \item \textbf{Fidelity to the Source Text}: Even when LLM outputs are structurally valid, numerically plausible, and semantically coherent, they may diverge from the original recipe through omissions, distortions, or oversimplification. Examples include omitting explicitly mentioned ingredients such as green chili, transforming `1.5 cups' into `1 cup', simplifying `chilled water' to `water', or extracting `cumin' instead of the more specific `roasted cumin powder'. Such deviations can erode the semantic, cultural, and functional fidelity of the resulting knowledge base. We address these issues using source-grounded verification methods that cross-check structured outputs against the original recipe text, including lexical and semantic parsing, retrieval-based evidence matching that links extracted facts to supporting spans in the source text, and fine-grained culinary Named Entity Recognition (NER) techniques designed for multilingual, code-switched, vernacular, and transliterated recipe content. These methods help identify invented, omitted, or misrepresented facts while maintaining scalability and reducing reliance on manual review.
    
    
\end{enumerate}

Many real-world errors span multiple categories. Examples observed during experimentation include confusing ingredient preparation states (e.g., \textit{birista} (deep-fried onions) versus raw onion), leading to incorrect texture, cooking time, and flavor; misinterpreting transliterated terms (e.g., \textit{saunf} (fennel seeds) versus soya (soy)) due to lexical overlap and phonetic ambiguity; applying substitutions that disregard regional culinary norms (e.g., replacing mustard oil with \textit{ghee} (clarified butter) in a Bengali fish curry); and misparsing code-switched instructions such as `add \textit{ek chamach} (one spoon) sugar,' where quantities, units, or ingredient boundaries may be incorrectly interpreted or omitted. 
The next section describes the multi-stage validation workflow. Although the validation architecture is not specific to Indian cuisine, several underlying resources require domain adaptation. Schema validation, numerical anomaly detection, source-grounded verification, and the overall detection-resolution architecture are broadly reusable, whereas ingredient vocabularies, co-occurrence distributions, substitution rules, and cultural compatibility models must be adapted to the target cuisine and language.

\section{Methodology}
\label{sec:methodology}

\subsection{Multi-Stage Soundness Assessment Workflow}
\label{subsec:soundness_assessment_workflow}

\begin{figure*}[t]
  \centering
  \includegraphics[
    width=\textwidth,
    trim=70 490 70 80,
    clip
  ]{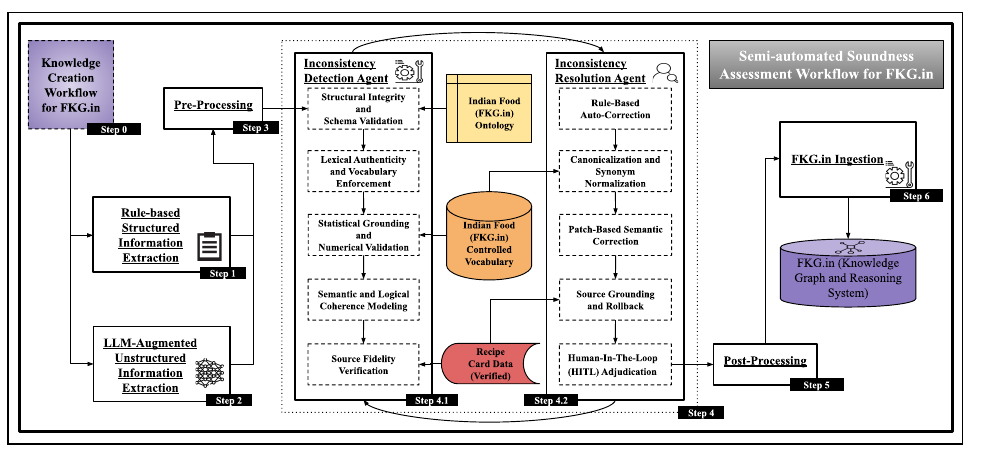}
  \caption{Semi-automated Soundness Assessment Workflow for FKG.in}
  \label{fig:soundness_assessment_workflow}
\end{figure*}

To address the challenges outlined in the previous section, we propose a multi-stage validation pipeline for FKG.in, a continuously evolving knowledge graph of Indian food \cite{Gupta2024FKG, Gupta2025FKG, Gupta2025extending}. The workflow combines LLM-based extraction with rule-based, statistical, and learning-based validation methods to identify and address structural, semantic, and source-fidelity issues before knowledge graph ingestion. Figure~\ref{fig:soundness_assessment_workflow} summarizes the overall workflow, which comprises seven sequential stages described below.


\begin{itemize}
    \item \textbf{Step 0: Knowledge Creation Workflow for FKG.in} --- FKG.in is a structured knowledge graph of Indian food constructed from both structured and unstructured culinary sources. In this work, we focus on recipe sources containing both structured recipe-card data and accompanying free-text recipe descriptions. The validation workflow focuses on structured information extracted from the free-text component, with LLM-based NER and subsequent augmentation performed prior to knowledge graph ingestion.
    
    \item \textbf{Step 1: Rule-based Structured Information Extraction} --- Structured recipe-card data are processed using deterministic parsers and schema-aligned templates to produce high-confidence structured representations. These extractions serve as reference data for benchmarking and validating information extracted by LLMs from the corresponding free-text recipe descriptions.

    \item \textbf{Step 2: LLM-Augmented Unstructured Information Extraction} --- LLMs are applied to free-text recipe descriptions to extract ingredients, quantities, preparation methods, and metadata. In addition to information attributes present in recipe cards, prompts are designed to elicit latent culinary attributes such as recipe provenance, preparation techniques, flavor characteristics, substitutions, and serving practices for additional context. All outputs are stored in a standardized JSON schema. For the extraction workflow used in this study, we employed OpenAI language models to generate structured representations from free-text recipe descriptions. The validation methods described in this work are independent of the underlying LLM and can be applied to outputs generated by other models. 

    \item \textbf{Step 3: Pre-Processing} --- LLM-extracted information is normalized into a canonical representation before validation. This stage performs character-level cleaning, noise filtering, synonym and substitution resolution, ingredient entity tagging, vernacular handling, and de-duplication. Extraneous punctuation, formatting artifacts, placeholders, instructions, and other non-ingredient content are removed using heuristic rules and a lightweight classifier. The extraction schema preserves synonym and substitution relationships. For example, an ingredient may be represented as `\textit{kala chana} (black chickpeas) or \textit{rajma}     (kidney beans)', where the canonical ingredient, synonym, and potential substitute are retained within a single representation and resolved into structured mappings. Ingredient names, quantities, units, and preparation states are extracted using a custom spaCy-based NER model \cite{honnibal2017spacy} trained on the FINER dataset \cite{Komariah2022} and further refined with manually curated ingredient annotations and `no entity' examples. Ingredients written in Devanagari are augmented with transliterated Hinglish representations to preserve linguistic variation while supporting semantic alignment. Finally, canonical ingredients are normalized through fuzzy matching (e.g., `tomato' vs. `tomatoes'), and metadata such as states, synonyms, and script variants are consolidated across duplicates.

    \item \textbf{Step 4: Soundness Assessment} --- This core stage identifies and addresses structural, numerical, semantic, and source-fidelity inconsistencies in LLM-extracted data. Validation methods include schema enforcement, anomaly detection, vocabulary-based filtering, contextual coherence modeling, and source-grounded verification. The individual inconsistency detection and resolution components are described in Subsections \ref{subsec:inconsistency_detection} and \ref{subsec:inconsistency_resolution}. Appendix \ref{appendix:implementation_status} summarizes the implementation and evaluation status of these components. These components operate iteratively, with corrected outputs re-evaluated until no critical inconsistencies remain or predefined confidence thresholds are satisfied, after which the data proceed to post-processing.

    \item \textbf{Step 5: Post-Processing} --- Following automated and human-assisted validation, the resulting data are consolidated into canonical ingredient representations by resolving residual ambiguity, unifying variants, and aligning entities with internal database identifiers. Semantic embeddings are generated for each ingredient using a sentence-transformer model to support similarity analysis, clustering, and coherence assessment, while source-fidelity metadata are retained to record deviations from the original textual mention. Near-duplicate ingredient expressions are clustered across recipes, and validated ingredient lists are aggregated to construct an ingredient co-occurrence matrix. Ingredient categories such as spices, legumes, and dairy are subsequently reintegrated into the canonical database to support structured querying and category-aware analysis. These steps enrich FKG.in and provide a semantically and statistically grounded baseline for future validations.

    \item \textbf{Step 6: Ingestion into FKG.in} --- Once validation is complete, structured data are serialized and incorporated into FKG.in. Relationships among ingredients, recipes, categories, substitutions, and related entities are materialized to support structured querying, reasoning, and downstream food computing applications.
\end{itemize}

\subsection{Inconsistency Detection Agent}
\label{subsec:inconsistency_detection}

We introduce an \textit{Inconsistency Detection Agent}, a multi-stage module that identifies structural, semantic, numerical, and source-fidelity inconsistencies using rule-based, statistical, and machine learning methods. Ingredient data are evaluated across all five detection stages, while other recipe attributes such as equipment, preparation time, and instructions are assessed only through schema validation and source-fidelity checks. These attributes exhibit less domain-specific ambiguity and cultural variability, are more reliably extracted by LLMs, and generally follow predictable lexical patterns, making intermediate semantic and numerical validation less critical.

\begin{itemize}
    \item \textbf{Stage 1: Structural Integrity and Schema Validation} --- We employ JSON Schema-based validation to ensure that extracted ingredients conform to a predefined data model. This stage flags malformed records, missing fields, and datatype mismatches. To accommodate variability in LLM outputs, we use a dual-tier validation strategy that distinguishes between \textit{essential attributes} (e.g., ingredients, serving size) and \textit{non-essential attributes} (e.g., etymology, texture). Missing foundational data triggers a critical error, while sparse non-essential fields are patched with a canonical `na' (not available) value. The system also enforces strict array-type constraints on complex attributes such as cooking techniques and kitchen tools, preventing structured data from being collapsed into flat, comma-separated strings.

    \item \textbf{Stage 2: Lexical Authenticity and Vocabulary Enforcement} --- This stage validates extracted ingredient names against a controlled vocabulary of over 23,000 canonical, synonymous, and vernacular ingredient terms spanning Devanagari and Hinglish representations. The vocabulary is derived from the ingredient lexicon produced during preprocessing and seeded from the FINER dataset \cite{Komariah2022}, supplemented with manually curated ingredient annotations and vernacular variants. We use \textit{RapidFuzz} string matching \cite{bachmann2025} with a near-exact similarity threshold to validate extracted names against the curated vocabulary, reducing false positives from spelling variations and phonetic drift. Ingredient mentions are also compared against corpus-level co-occurrence statistics to identify improbable substitutions and anomalous insertions.
        
    \item \textbf{Stage 3: Statistical Grounding and Numerical Validation} --- This stage detects improbable numerical values by comparing ingredient quantities to distributional norms across the corpus, including serving size--normalized comparisons. To identify anomalies such as `5 kg of \textit{dal} for 1 serving,' the system applies adaptive bounds based on the Interquartile Range (IQR) of the weight-per-unit distribution for each ingredient-unit pair. For ingredient-unit pairs with limited historical support ($n < 3$ observations), a low-data heuristic expands the bounds by 40\% to prevent over-flagging valid but uncommon measurements. Before statistical validation, ingredient strings are normalized using the custom spaCy-based NER pipeline to ensure accurate lookup against the distributional model.
    
    \item \textbf{Stage 4: Semantic and Logical Coherence Modeling} --- We employ a 12-layer Set Transformer \cite{lee2019set} with 16 attention heads and a 1536-dimensional hidden layer to assess the contextual plausibility of ingredient sets. The model uses 1024-dimensional \textit{GTE-Large} embeddings \cite{li2023towards} to compute the semantic fit of each ingredient within the recipe context. Rather than using a fixed probability threshold, the system calculates an adaptive IQR upper bound on the \textit{discordance} probabilities for the current set, allowing the agent to flag the most semantically discordant items---such as tuna in \textit{paneer tikka} or \textit{maida} in \textit{kadhi}---relative to the specific recipe distribution and identify illogical combinations that may violate regional or cultural culinary norms.

        \item \textbf{Stage 5: Source Fidelity Verification (Dual-Judge SLMs)} --- We utilize a vectorized, batch-optimized dual-judge system based on small language models (SLMs), specifically 435M-parameter \textit{DeBERTa-v3-large} models \cite{he2021debertav3}, to verify extracted facts against the source text. The system comprises two specialized agents: (1) a \textit{Relation Extraction (RE) Judge} that implements mutual exclusivity logic, selecting the globally best non-overlapping set of ingredient-quantity pairs based on confidence scores, and (2) a \textit{Natural Language Inference (NLI) Reasoner} that maps \textit{Neutral} predictions to \textit{Contradiction} to provide a conservative validation baseline. To account for implicit culinary conventions, a \textit{Staple Guard} automatically exempts ambient ingredients (e.g., salt, water, oil) from fidelity rejections. This stage identifies omissions and factual drift that surface-level heuristics may miss.
\end{itemize}

\subsection{Inconsistency Resolution Agent}
\label{subsec:inconsistency_resolution}

While the detection pipeline identifies diverse failure modes, detection alone is insufficient. We therefore present a multi-stage Inconsistency Resolution Agent that repairs, normalizes, or escalates inconsistencies according to their severity and recoverability, progressing from deterministic rule-based repairs and canonicalization to patch-based semantic correction, source-grounded verification, and Human-in-the-Loop (HITL) adjudication.

\begin{figure*}[t]
  \centering
  \includegraphics[width=\textwidth]{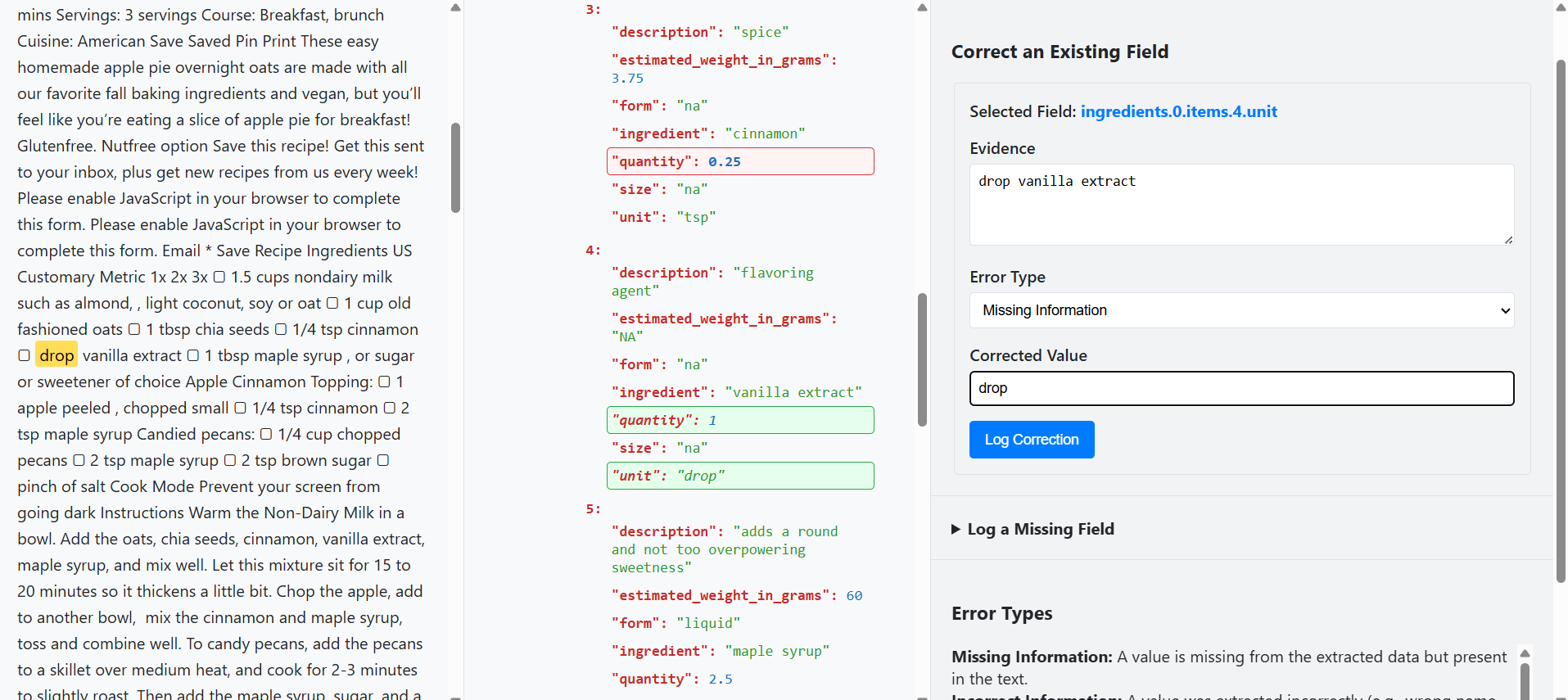}
  \caption{Annotator Interface for Human-in-the-Loop Verification and Resolution} 
  \label{fig:human_in_the_loop}
\end{figure*}

\begin{itemize}
    \item \textbf{Stage 1: Rule-Based Auto-Correction} --- This stage addresses deterministic, surface-level inconsistencies flagged during schema validation using handcrafted rules, regex templates, and culinary syntax norms. The system performs lightweight rewrites to improve structural conformance. For example, malformed expressions such as `200mlgram' are disambiguated into `200 ml' or `200 grams' using ingredient-specific unit priors (oil resolves to ml, flour to grams). Similarly, `salt 1 tsp' is reordered as `1 tsp salt', and `onions - chopped' becomes `chopped onions' to align with schema conventions.

    \item \textbf{Stage 2: Canonicalization and Synonym Normalization} --- After Vocabulary Enforcement, ingredient names are aligned to a controlled vocabulary using fuzzy matching, phonetic encoding, and co-occurrence-based clustering. Variant expressions are mapped to canonical forms to reduce duplication. For example, \textit{haldi}, turmeric powder, and ground turmeric are normalized to turmeric; capsicum and bell pepper are unified; and verbose modifiers (e.g., finely chopped fresh coriander leaves) are reduced to their semantic core, i.e., fresh coriander.

    \item \textbf{Stage 3: Patch-Based Semantic Correction} --- This stage addresses complex semantic contradictions and recurring extraction errors through case-based reuse of historical human corrections. When an inconsistency is detected, the agent performs a K-Nearest Neighbor search over a repository of human-validated patches. If a similar error pattern is identified (similarity $>$ 0.85), a secondary model (Gemini 1.5 Flash) adapts the validated correction to the current recipe context rather than generating one from scratch. For example, a patch validated for \textit{Jain} recipes may be reused when an extracted ingredient list incorrectly contains egg, replacing it with a contextually appropriate alternative such as \textit{paneer}; lexical duplications such as gram flour and \textit{besan} can be consolidated into a single canonical entry; and logically incompatible ingredient combinations, such as milk and tofu in non-fusion contexts, can be pruned or flagged for review. The module uses dish-level metadata (e.g., cuisine, course, and dietary tags), dietary substitution graphs, constraint-aware templates, and historical correction patterns to apply conservative, context-aware edits.

    \item \textbf{Stage 4: Source Grounding and Rollback} --- To resolve hallucinations or factual drift flagged during source fidelity checks, this module performs a targeted verification pass against the original recipe source. Retrieval techniques re-examine relevant text spans to ground each ingredient or attribute (e.g., quantity, unit, preparation method) in source evidence. If a suspected hallucination lacks a verifiable source anchor and cannot be reasonably inferred from culinary context, such as an ingredient implicitly included in \textit{garam masala} (a blended spice mixture), the unsupported node is rolled back. For example, if clove appears in the extracted ingredient list but cannot be found in the source recipe, it is flagged for removal.
    
    \item \textbf{Stage 5: Human-in-the-Loop (HITL) Adjudication} --- As the final fallback, this stage handles complex or ambiguous cases where automated resolution is insufficient or risky, including missed entries, region-specific substitutions, overlapping ingredient categories, culturally sensitive dishes, and uncertain fusion contexts. Examples include determining whether \textit{garam masala} and its constituent spices are redundant, distinguishing plantain from raw banana, or resolving co-occurrences such as \textit{paneer} and tofu. A web-based annotator interface supports accept/reject/suggest workflows and provides relevant source evidence (Figure~\ref{fig:human_in_the_loop}). Human decisions are logged to expand the correction database and refine substitution graphs, co-occurrence priors, and canonicalization rules, strengthening the Stage 3 Patch-Based Correction module as more data are adjudicated.
\end{itemize}

A proof-of-concept video illustrating the HITL validation and resolution workflow is included in the supplementary material. The soundness assessment pipeline and the HITL correction tool are available on GitHub\footnote{\url{https://github.com/fkg-india/validating-fkg.in}} and 
archived on Zenodo for long-term access~\cite{fcl_ashoka}.

\section{Results}
\label{sec:results}

We evaluated the detection pipeline on 500 recipe documents processed using GPT-based extraction from free-text culinary sources and separately evaluated the patch-based resolution component on previously unseen contradictions identified during source-fidelity verification. Rather than measuring extraction performance alone, we characterized the prevalence and distribution of LLM-induced inconsistencies relevant to downstream knowledge graph construction. Failures were observed across all validation dimensions, ranging from schema violations and numerical anomalies to semantically implausible ingredient combinations and contradictions with the source text.

\subsection{Evaluation of the Inconsistency Detection Agent}

Table~\ref{tab:pipeline_performance} summarizes the proportion of documents affected and the number of instances identified by the Inconsistency Detection Agent across stages, with detailed stage-wise results provided in Appendix~\ref{appendix:error_detection_performance}.

\begin{table}
  \caption{Failure rates and primary failure modes identified by the Inconsistency Detection Agent (N=500).}
  \label{tab:pipeline_performance}
  \centering
  \begin{tabular}{
    @{}
    >{\raggedright\arraybackslash}p{0.23\columnwidth}
    >{\raggedright\arraybackslash}p{0.20\columnwidth}
    >{\raggedright\arraybackslash}p{0.13\columnwidth}
    >{\raggedright\arraybackslash}p{0.40\columnwidth}
    @{}
  }
    \toprule
    \textbf{Validation Stage} &
    \textbf{Affected Docs.} &
    \textbf{Instances} &
    \textbf{Representative Failure Mode} \\
    \midrule

    Detection Stage 1 &
    91.6\% (458) &
    1,436 &
    Invalid data types in sparse arrays \\

    Detection Stage 2 &
    48.0\% (240) &
    544 &
    Unmapped or unfamiliar ingredient expressions \\

    Detection Stage 3 &
    36.6\% (183) &
    267 &
    Numerically implausible ingredient quantities \\

    Detection Stage 4 &
    68.8\% (344) &
    605 &
    Contextually anomalous ingredients \\

    Detection Stage 5 &
    96.4\% (482) &
    2,447 &
    Contradictions with source text \\

    \bottomrule
  \end{tabular}
\end{table}

The distribution of failures across stages highlights the limitations of relying on any single validation mechanism. Structural violations occurred in 91.6\% of documents, while vocabulary enforcement, statistical grounding, and semantic coherence modeling identified errors in 48.0\%, 36.6\%, and 68.8\% of documents, respectively. These errors would largely remain undetected through schema validation alone. Most notably, source-fidelity verification identified contradictions in 96.4\% of documents, indicating that structurally valid and semantically plausible extractions may still diverge substantially from the source text. The 48.0\% failure rate observed during vocabulary enforcement highlights a persistent vernacular gap. LLMs frequently preserve source terminology without aligning it to canonical ingredient representations, reinforcing the need for the lexical authenticity and vocabulary enforcement mechanisms described in Subsection~\ref{subsec:inconsistency_detection}.

Analysis of over 2,000 source-fidelity errors identified by Stage 5 revealed three dominant classes of extraction failure. Quantitative hallucinations accounted for 49.0\% of fidelity errors (1,201 instances), where the correct ingredient was extracted but paired with an unsupported quantity or unit. Existential hallucinations represented 29.7\% (728 instances), reflecting the tendency of LLMs to introduce assumed ingredients, such as common culinary staples, that were absent from the source text. Morphological contradictions accounted for 4.0\% (100 instances), typically involving incorrect preparation states such as extracting `powder' when the source specified `whole.' Together, these findings suggest that many ingredient-level extraction failures arise not from entity recognition itself, but from semantic drift and unsupported inference during structured generation.

\subsection{Evaluation of the Inconsistency Resolution Agent}

To evaluate the inconsistency resolution workflow, we conducted a strict cross-recipe evaluation using 1,494 previously unseen contradictions identified during Stage 5 (Source Fidelity Verification). The correction repository was initialized with only 42 human-validated corrections. Retrieval was performed using a weighted two-tower embedding strategy based on GTE-Large, assigning 80\% weight to the hallucinated entity and 20\% weight to recipe context in order to reduce context washout effects.

The retrieval mechanism achieved 53.48\% coverage of previously unseen contradictions at a cosine similarity threshold of $\geq 0.88$, indicating that many extraction failures recur in structurally similar forms across recipes. To preserve knowledge graph integrity, automated correction was restricted to high-confidence interventions consisting only of exact entity matches and staple/ambient inferences. Under these conservative constraints, the system autonomously resolved 311 contradictions, corresponding to 20.8\% coverage of all unseen errors.

Although intentionally conservative, these results demonstrate that a relatively small repository of validated corrections can be reused to resolve a substantially larger number of recurring extraction failures. In this setting, 42 human-validated examples yielded 311 autonomous corrections, reducing manual review effort while maintaining strict quality controls. The findings suggest that many LLM extraction errors follow recurring patterns that can be captured through human validation and subsequently reused across recipes, supporting the long-term scalability of the proposed resolution workflow.

\subsection{Ingredient NER Evaluation}

Because lexical authenticity checks depend on accurate ingredient recognition, we additionally evaluated the ingredient NER component used during preprocessing and vocabulary construction. The evaluation suites were intentionally designed to differ from the model's training distribution and to stress linguistic phenomena relevant to recipe extraction, including ingredient coordination, vernacular terminology, and complex ingredient descriptions.

Table~\ref{tab:ner_performance} summarizes performance across four evaluation suites. Precision, recall, and F$_1$ scores were computed at the entity level, requiring exact agreement on both text span boundaries and labels. These evaluation suites serve as diagnostic stress tests intended to assess robustness under realistic extraction conditions rather than benchmark performance.

\begin{table}
  \caption{Ingredient NER performance on evaluation suites assessing generalization beyond training distribution.}
  \label{tab:ner_performance}
  \centering
  \begin{tabular}{
    @{}
    >{\raggedright\arraybackslash}p{0.23\columnwidth}
    >{\centering\arraybackslash}p{0.20\columnwidth}
    >{\centering\arraybackslash}p{0.15\columnwidth}
    >{\centering\arraybackslash}p{0.13\columnwidth}
    >{\raggedright\arraybackslash}p{0.26\columnwidth}
    @{}
  }
    \toprule
    \textbf{Evaluation Suite} &
    \textbf{Precision (\%)} &
    \textbf{Recall (\%)} &
    \textbf{F$_1$ (\%)} &
    \textbf{Example(s)} \\
    \midrule

    Simple Ingredient Phrases &
    82.76 &
    85.71 &
    84.21 &
    `2 cups of flour'; `salt' \\

    Coordinated Multi-Ingredient Sentences &
    51.16 &
    49.44 &
    50.29 &
    `2 cups of flour and 1 tsp yeast' \\

    Hinglish and Vernacular Expressions &
    65.12 &
    65.12 &
    65.12 &
    `2 \textit{chamach} (spoon) ghee'; `\textit{haldi} powder' \\

    Complex Ingredient Descriptions &
    71.05 &
    79.41 &
    75.00 &
    `whole San Marzano tomatoes, with juices' \\

    \bottomrule
  \end{tabular}
\end{table}

Performance remained strong on simple ingredient expressions (F$_1$ = 84.2\%) and reasonably robust on complex constructions (F$_1$ = 75.0\%), while longer compositional sentences (F$_1$ = 50.3\%) and Hinglish or vernacular expressions (F$_1$ = 65.1\%) proved more challenging. The results suggest that ambiguity introduced by coordination, boundary detection, and regional terminology remains a source of extraction error. These findings are consistent with the motivation for the downstream vocabulary-enforcement, semantic validation, and human-review stages, which are designed to compensate for residual extraction ambiguity and linguistic variability.

Taken together, the results show that LLM-extracted culinary knowledge exhibits failure modes spanning structural validity, lexical normalization, numerical plausibility, semantic coherence, and source fidelity. The prevalence of errors detected across all validation stages indicates that no single validation mechanism is sufficient for reliable knowledge graph ingestion. These findings provide empirical support for the layered design of the proposed soundness assessment workflow, while the retrieval-based resolution component demonstrates a practical path toward reducing manual curation effort as validated corrections accumulate.

A detailed proof-of-concept walkthrough of a representative \textit{dum aloo} (slow-cooked potato curry) recipe, illustrating stage-wise detection and subsequent resolution, is provided in Appendix~\ref{appendix:poc_detection}.

\section{Limitations and Future Directions}
\label{sec:limitations}

Systematic evaluation beyond the patch-based resolution component remains ongoing. Semantically ambiguous, culturally nuanced, or region-specific culinary concepts require human oversight in some cases. Although the Inconsistency Detection and Resolution Agents are designed as complementary modules, their feedback loop is not yet fully integrated, and corrections generated during resolution are not systematically leveraged to improve subsequent detection. In addition, the current deployment of the Inconsistency Resolution Agent relies on conservative similarity thresholds ($\geq 0.88$) and strict error archetypes (e.g., exact entity matches) to prevent automation drift. While this design prioritizes knowledge graph integrity, it limits autonomous handling of novel or highly complex semantic inconsistencies without human-validated correction seeds. The current framework also focuses primarily on text-based recipe sources; multimodal content such as video recipes or image-centric blogs will require more sophisticated alignment and grounding. Detection rates indicate the proportion of recipe documents flagged at each stage, not independently verified precision or recall.

Future work will focus on systematic stage-wise evaluation of the detection and resolution components using expanded and curated benchmark datasets, enabling finer-grained assessment of structural validity, semantic coherence, source fidelity, and resolution quality. We also plan to strengthen the interaction between detection and resolution through iterative feedback mechanisms, extend the framework to multilingual recipe sources beyond English and Devanagari, improve contextual reasoning through retrieval-augmented methods, and expand support for culturally nuanced ingredient substitutions and preparation practices. As FKG.in evolves, these efforts aim to reduce manual review while preserving the transparency, traceability, and culinary fidelity required for trustworthy food knowledge curation. Beyond FKG.in, the lessons learned may inform soundness assessment workflows for other domain-specific knowledge graphs built using LLM-assisted extraction.

\section{Conclusion}
\label{sec:conclusion}

As large language models become increasingly central to knowledge extraction pipelines, ensuring the reliability of their outputs is essential for building trustworthy domain knowledge graphs. In this work, we presented a semi-automated soundness assessment workflow for FKG.in, a continuously evolving knowledge graph of Indian food, designed to identify structural inconsistencies, numerical anomalies, semantic incoherence, and source-fidelity violations in LLM-extracted recipe data. Evaluation across 500 recipe documents demonstrated that extraction failures occur across all validation dimensions, with many remaining undetectable through schema validation alone, underscoring the need for layered verification before knowledge graph ingestion. By combining schema validation, statistical grounding, lexical and semantic consistency checks, source-grounded verification, retrieval-assisted correction, and human-in-the-loop review, the workflow provides multiple layers of quality control and a path toward reducing manual curation effort. Although motivated by Indian culinary data, the framework offers a practical template for validating LLM-generated structured knowledge in other domains where extracted information must remain faithful, coherent, traceable, and auditable.

\section{Acknowledgments}
This research was supported by the Mphasis AI \& Applied Tech Lab at Ashoka, a collaboration between Ashoka University and Mphasis Limited, and the Koita Centre for Digital Health at Ashoka University.

%
%
\bibliographystyle{splncs04}
\bibliography{main}

\appendix

\section{Inconsistency Detection and Resolution: Proof of Concept on a Sample Recipe}
\label{appendix:poc_detection}

To demonstrate the operation of the proposed validation workflow, we present a walkthrough of both the \textbf{Inconsistency Detection Agent} and the \textbf{Inconsistency Resolution Agent}. For this purpose, we selected a representative \textit{dum aloo} (slow-
cooked potato curry) recipe document (a raw long-form recipe blog text describing the recipe) from the evaluation corpus and extracted the corresponding LLM-generated structured JSON representation from the raw recipe text. The initial extraction was then manually annotated to identify representative failure modes and assess how they are handled by the validation pipeline.

We identified the following errors in the raw LLM extraction:
\begin{itemize}
    \item \textbf{Incorrect Data (Factual Drift \& Morphological Contradictions)}
    \begin{itemize}
        \item \textbf{Error (1):} baby potatoes: quantity reported as 12.5; actual is 13.5.
        \item \textbf{Error (2):} baby potatoes: form reported as `whole'; actual is `boiled, peeled, and pricked'.
        \item \textbf{Error (3):} oil: quantity reported as 5; actual is 2.5.
        \item \textbf{Error (4):} coriander seeds: form reported as \texttt{na} \footnote{lowercased, cleaned, and abbreviated N/A or 'Not Available'}; actual is `crushed'.
        \item \textbf{Error (5):} onion: quantity reported as 3; actual is 1.
    \end{itemize}

    \item \textbf{Hallucination (Existential)}
    \begin{itemize}
        \item \textbf{Error (6):} Serving size reported as 6, but not present in the source.
    \end{itemize}
\end{itemize}
We also added two inconsistent ingredients to the extracted JSON to further demonstrate the coherence capabilities of the framework. The stage-by-stage demonstration is as follows:

\subsection{Inconsistency Detection Agent Walkthrough}

\begin{itemize}
    \item \textbf{Stage 1: Structural Integrity and Schema Validation} --- The structure of the extracted JSON is validated against the expected schema. Several required fields were missing in the initial output: \texttt{place\_of\_origin}, \texttt{ferment\_time}, \texttt{knife\_cuts}, \texttt{flavor}, \texttt{texture}, \texttt{taste}, and \texttt{related\_recipes}. These omissions were automatically flagged.

    \item \textbf{Stage 2: Lexical Authenticity and Vocabulary Enforcement} --- A controlled vocabulary check is applied to ensure all ingredients conform to a predefined list. The manually inserted ingredient \textit{dum aloo} (added for demonstration and representing the dish itself rather than an ingredient) was flagged by this stage, illustrating a failure mode in which the LLM incorrectly classifies a recipe name as an ingredient.

    \item \textbf{Stage 3: Statistical Grounding and Numerical Validation} --- Statistical analysis is used to identify ingredient weights that significantly deviate from historical ingredient-unit distributions. Several ingredients, including ginger garlic paste, coriander powder, and cumin powder, were flagged as numerical outliers and routed for review. These anomalies may indicate quantity extraction errors or implausible LLM-generated weight estimates.

    \item \textbf{Stage 4: Semantic and Logical Coherence Modeling} --- The Set Transformer assesses how well each ingredient aligns with the overall recipe context. Two ingredients, \textit{dum aloo} and bread (both manually added for demonstration), were identified as semantic outliers with discard probabilities of 34\% and 54\%, respectively. In contrast, valid ingredients exhibited probabilities near 0.1\%, indicating that the injected ingredients were clear contextual deviations.

    \item \textbf{Stage 5: Source Fidelity Verification} --- The Dual-Judge SLM system evaluates the alignment between the structured output and the source text. 
    \begin{itemize}
        \item It flagged \textbf{Error (2)} (`whole' vs `boiled, peeled, and pricked') as a \textit{Morphological Contradiction}.
        \item It flagged \textbf{Error (6)} (serving size: 6) as an \textit{Existential Hallucination} because no numerical serving size was anchored in the source.
        \item It correctly mapped the extracted ingredient components against the textual bounds, successfully identifying factual drift in quantitative values (Errors 1, 3, and 5).
    \end{itemize}
\end{itemize}

\subsection{Inconsistency Resolution Agent Walkthrough}

Once the detection agent flags the inconsistencies, the \textbf{Inconsistency Resolution Agent} systematically processes the flagged nodes using its tiered correction loop. Not all resolution stages were activated for this example; only stages relevant to the detected inconsistencies are illustrated below.

\begin{itemize}
    \item \textbf{Stage 1: Rule-Based Auto-Correction} --- The system addresses the missing sparse fields flagged in Detection Stage 1. It applies a schema-driven repair by patching missing non-essential fields (e.g., \texttt{knife\_cuts}, \texttt{ferment\_time}) with the canonical \texttt{na} value, ensuring database interoperability without invoking the LLM.
    
    \item \textbf{Stage 3: Patch-Based Semantic Correction} --- To resolve the Morphological Contradiction in \textbf{Error (2)} (where baby potatoes were marked as `whole'), the system embeds the contextual error using the Weighted Two-Tower strategy. It performs a K-Nearest Neighbor (KNN) search across the seed database of 42 human corrections. 
    
    It retrieves a highly similar historical patch (Cosine Similarity $\geq 0.88$): \textit{`If the source text dictates specific physical preparation states for an entity, extract them as modifiers to the form/size attribute and replace the LLM's default assumed state.'} 
    
    A lightweight secondary model (Gemini 1.5 Flash) applies this retrieved logic to the local extraction, successfully replacing `whole' with `boiled, peeled, and pricked' without manual intervention.

    \item \textbf{Stage 4: Source Grounding and Rollback} --- To resolve the Existential Hallucination in \textbf{Error (6)} (serving size: 6), the system attempts a targeted retrieval pass to locate the span `6' in the source text adjacent to serving-related vocabulary. Upon failing to find a supporting anchor, the system executes a deterministic \textit{rollback}, removing the hallucinated \texttt{serving\_size} key-value pair entirely.
\end{itemize}

Through this closed-loop architecture, the system can resolve structural omissions, semantic contradictions, and existential hallucinations. Complex numerical range errors (e.g., Errors 1 and 3) that fail to meet the strict $\geq 0.88$ similarity threshold for autonomous patching are escalated to the Human-in-the-Loop (HITL) interface for review, helping preserve the knowledge graph's integrity.

\section{Error Detection Performance (N=500)}
\label{appendix:error_detection_performance}

Below is a summary of the Inconsistency Detection Agent's results across the 500-recipe corpus. Table~\ref{tab:early_detection_summary} summarizes the first four detection stages: Stage 1 identified structural violations, including invalid data types in sparse arrays; Stage 2 identified unmapped or unfamiliar ingredient expressions, including vernacular variants and synonyms; Stage 3 detected numerical outliers, particularly implausible weight-to-unit ratios; and Stage 4 identified contextually anomalous ingredients. Table~\ref{tab:error_categories} summarizes the source-fidelity errors identified by Stage 5, including ingredient-level contradictions and metadata discrepancies between the extracted information and the source text.

\begin{table}[!htbp]
  \centering
  \small
    \caption{Detailed quantitative results for detection Stages 1--4.}
  \label{tab:early_detection_summary}
  \begin{tabular}{
    @{}
    >{\centering\arraybackslash}p{0.07\columnwidth}
    >{\raggedright\arraybackslash}p{0.26\columnwidth}
    >{\centering\arraybackslash}p{0.13\columnwidth}
    >{\centering\arraybackslash}p{0.12\columnwidth}
    >{\centering\arraybackslash}p{0.17\columnwidth}
    >{\centering\arraybackslash}p{0.17\columnwidth}
    @{}
  }
    \toprule
    \textbf{Stage} &
    \textbf{Detection measure} &
    \textbf{Instances} &
    \textbf{Documents} &
    \textbf{Affected (\%)} &
    \textbf{Mean / doc.} \\
    \midrule
    1 & Structural violations &
    1,436 & 458 & 91.6 & 2.87 \\
    2 & Unmapped or unfamiliar ingredients &
    544 & 240 & 48.0 & 1.09 \\
    3 & Numerical outliers &
    267 & 183 & 36.6 & 0.53 \\
    4 & Semantic outliers &
    605 & 344 & 68.8 & 1.21 \\
    \bottomrule
  \end{tabular}
\end{table}

\begin{table}[!htbp]
  \caption{Distribution of fidelity errors by category.}
  \label{tab:error_categories}
  \begin{tabular}{
    @{}
    >{\raggedright\arraybackslash}p{0.68\columnwidth}
    >{\centering\arraybackslash}p{0.13\columnwidth}
    >{\centering\arraybackslash}p{0.13\columnwidth}
    @{}
  }
    \toprule
    \textbf{Error Category} &
    \textbf{Total Errors} &
    \textbf{Share (\%)} \\
    \midrule
    \texttt{ingredient\_fidelity - existence\_contradictions} & 728 & 29.75 \\ 
    \texttt{ingredient\_fidelity - quantity\_contradictions} & 1,201 & 49.08 \\ 
    \texttt{ingredient\_fidelity - form\_contradictions} & 100 & 4.09 \\ 
    \texttt{metadata - metadata\_field\_not\_in\_extracted: prep\_time} & 21 & 0.86 \\ 
    \texttt{metadata - metadata\_field\_not\_in\_extracted: cook\_time} & 24 & 0.98 \\ 
    \texttt{metadata - metadata\_field\_not\_in\_source: serving\_size} & 78 & 3.19 \\ 
    \texttt{metadata - metadata\_field\_not\_in\_source: prep\_time} & 58 & 2.37 \\ 
    \texttt{metadata - metadata\_field\_not\_in\_extracted: serving\_size} & 73 & 2.98 \\ 
    \texttt{metadata - metadata\_field\_not\_in\_source: total\_time} & 86 & 3.51 \\ 
    \texttt{metadata - metadata\_field\_not\_in\_extracted: total\_time} & 27 & 1.10 \\ 
    \texttt{metadata - metadata\_field\_not\_in\_source: cook\_time} & 41 & 1.68 \\ 
    \texttt{metadata - metadata\_value\_mismatch: serving\_size} & 4 & 0.16 \\ 
    \texttt{metadata - metadata\_value\_mismatch: prep\_time} & 1 & 0.04 \\ 
    \texttt{metadata - metadata\_value\_mismatch: cook\_time} & 2 & 0.08 \\ 
    \texttt{metadata - metadata\_value\_mismatch: total\_time} & 3 & 0.12 \\ 
  \bottomrule
\end{tabular}
\end{table}

\section{Implementation Status}
\label{appendix:implementation_status}

Table~\ref{tab:implementation_status} provides a component-wise account of the implementation and evaluation progress for the Inconsistency Detection and Inconsistency Resolution Agents.

\begin{table}
  \caption{Implementation and evaluation status of the proposed detection and resolution stages.}
  \label{tab:implementation_status}
  \begin{tabular}{
    @{}
    >{\raggedright\arraybackslash}p{0.65\columnwidth}
    >{\raggedright\arraybackslash}p{0.33\columnwidth}
    @{}
  }
    \toprule
    \textbf{Component} & \textbf{Status} \\
    \midrule
    Detection 1: Structural Integrity and Schema Validation &
    Implemented; evaluated \\

    Detection 2: Lexical Authenticity and Vocabulary Enforcement &
    Implemented; evaluated \\

    Detection 3: Statistical Grounding and Numerical Validation &
    Implemented; evaluated \\

    Detection 4: Semantic and Logical Coherence Modeling &
    Implemented; evaluated \\

    Detection 5: Source Fidelity Verification &
    Implemented; evaluated \\

    Resolution 1: Rule-Based Auto-Correction &
    Implemented; demonstrated \\

    Resolution 2: Canonicalization and Synonym Normalization &
    Implemented \\

    Resolution 3: Patch-Based Semantic Correction &
    Implemented; evaluated \\

    Resolution 4: Source Grounding and Rollback &
    Implemented; demonstrated \\

    Resolution 5: Human-in-the-Loop Adjudication &
    Implemented; demonstrated \\
    \bottomrule
  \end{tabular}
\end{table}

\end{document}